%% file: egpaper_for_review.tex
\documentclass[runningheads]{llncs}

\usepackage{epsfig}
\usepackage{graphicx}
\usepackage{amsmath}
\usepackage{amssymb}
\usepackage{paralist}
\usepackage{lipsum}
\usepackage{enumitem}
\usepackage{subcaption}
\usepackage{balance}
\usepackage{xcolor}
\usepackage{makecell}
\usepackage{caption} 
\usepackage{color, colortbl}

\definecolor{Gray}{gray}{0.9}

\definecolor{LightCyan}{rgb}{0.88,1,1}
\usepackage{hyperref}

\usepackage[width=122mm,left=12mm,paperwidth=146mm,height=193mm,top=12mm,paperheight=217mm]{geometry}

\newcommand\blfootnote[1]{%
  \begingroup
  \renewcommand\thefootnote{}\footnote{#1}%
  \addtocounter{footnote}{-1}%
  \endgroup
}

\newcommand{\Autoref}[1]{%
  \begingroup%
  \def\chapterautorefname{Chapter}%
  \def\sectionautorefname{Section}%
  \def\subsectionautorefname{Subsection}%
  \autoref{#1}%
  \endgroup%
}

\begin{document}

\pagestyle{headings}

\mainmatter
\def\ECCVSubNumber{6927}  

\title{Learning Permutation Invariant Representations using Memory Networks} 
 

\titlerunning{Memory-based Exchangeable Model (MEM)}
\author{Shivam Kalra$^*$\inst{1}, 
  Mohammed Adnan$^*$\inst{1,2}, 
  Graham Taylor\inst{2,3},  H.R. Tizhoosh\inst{1,2}}
\authorrunning{S. Kalra et al.}
%
\institute{Kimia Lab, University of Waterloo, Canada \and
  Vector Institute for Artificial Intelligence, MaRS Centre, Toronto, Canada \and School of Engineering, University of Guelph, Canada\\
  \email{\{shivam.kalra,m7adnan\}@uwaterloo.ca \quad gwtaylor@uoquelph.ca \quad
    tizhoosh@uwaterloo.ca}}
\maketitle

\begin{abstract}
\blfootnote{$^*$Authors have equal contribution.} Many real world tasks such as
classification of digital histopathological images and 3D object detection involve
learning from a set of instances. In these cases, only a group of instances or a
set, collectively, contains meaningful information and therefore only the sets
have labels, and not individual data instances. In this work, we present a
permutation invariant neural network called \emph{Memory-based Exchangeable
Model (MEM)} for learning universal set functions. The MEM model consists of memory units
which embed an input sequence to high-level features enabling it to learn
inter-dependencies among instances through a self-attention mechanism. We
evaluated the learning ability of MEM on various toy datasets, point cloud
classification, and classification of whole slide images (WSIs) into two
subtypes of lung cancer---Lung Adenocarcinoma, and Lung Squamous Cell Carcinoma.
We systematically extracted patches from WSIs of lung, downloaded from The Cancer
Genome Atlas~(TCGA) dataset, the largest public repository of WSIs, achieving a
competitive accuracy of 84.84\% for classification of two sub-types of lung
cancer. The results on other datasets are promising as well, and demonstrate the
efficacy of our model. 

\end{abstract}

\section{Introduction}
\label{sec:intro}

\begin{figure}[t]
  \centering
  \begin{subfigure}[b]{0.7\textwidth}
    \centering
    \includegraphics[width=0.9\textwidth]{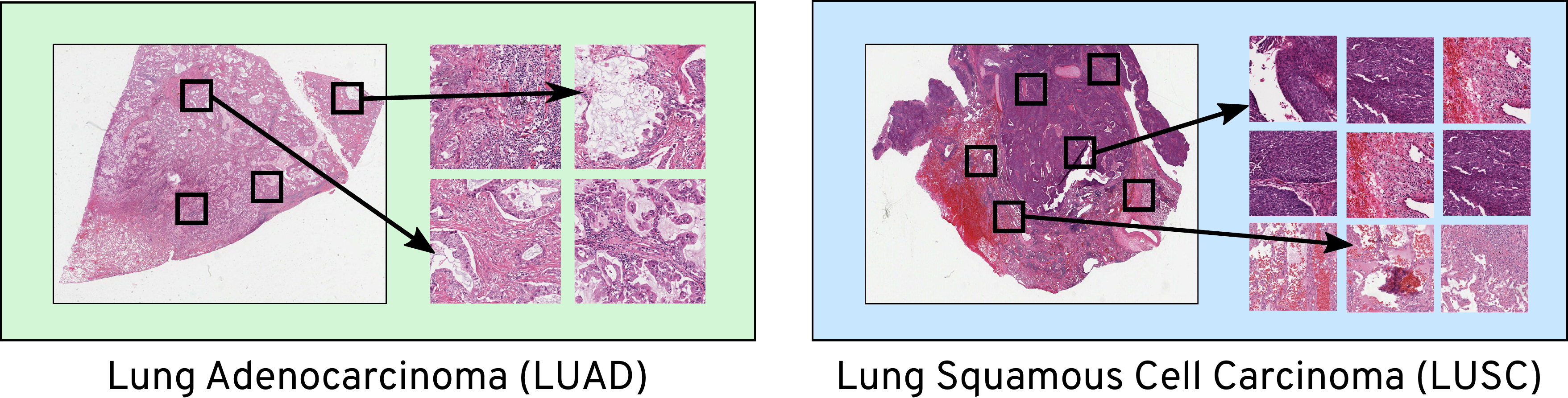}    
    \caption{Patch Extraction}
    \label{fig:intro-fig1}
  \end{subfigure}
  \begin{subfigure}[b]{0.7\textwidth}
    \centering
    \includegraphics[width=0.9\textwidth]{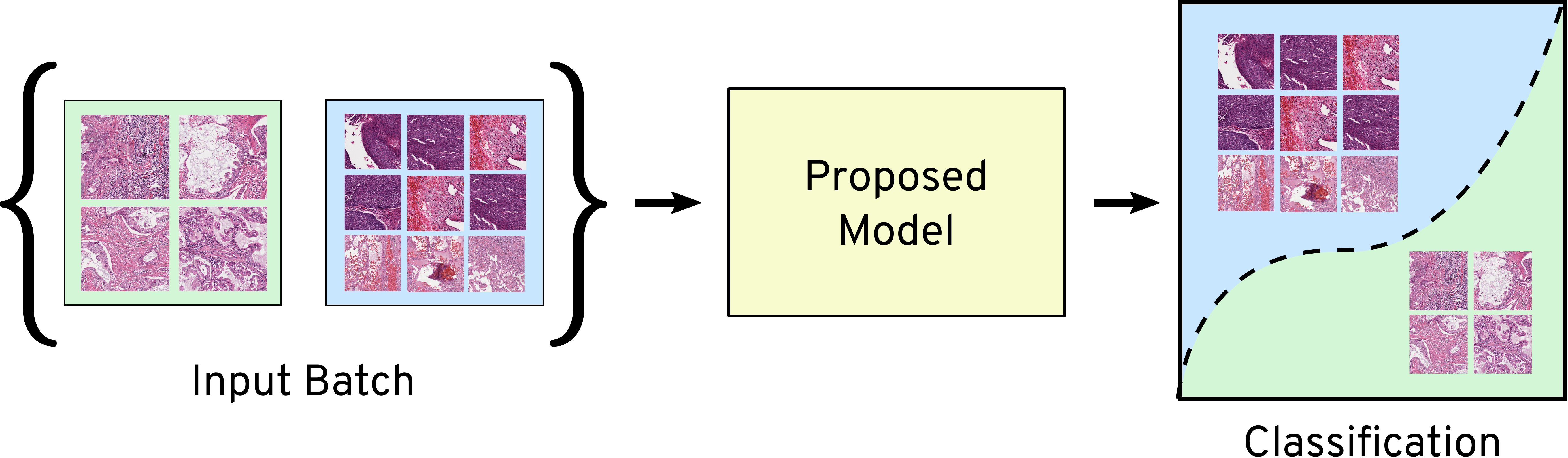}
    \caption{Set Classification}
    \label{fig:intro-fig2}
  \end{subfigure}
  \caption{An exemplar application of learning permutation invariant
    representation for disease classification of Whole-Slide Images (WSIs). (a) A
    set of patches are extracted from each WSI of patients with lung cancer. (b)
    The sets of patches are fed to the proposed model for classification of the
    sub-type of lung cancer---LUAD versus LUSC. The model classifies on a per set
    basis. This form of learning is known as Multi Instance Learning (MIL).}
    \label{fig:intro-luadlusc}
\end{figure}

Deep artificial neural networks have achieved impressive performance for
representation learning tasks. The majority of these deep architectures take a
single instance as an input. Recurrent Neural Networks (RNNs) are a popular
approach to learn representations from sequential ordered instances. However,
the lack of permutation invariance renders RNNs ineffective for exchangeable or
unordered sequences. We often need to learn representations of {unordered}
sequential data, or {exchangeable sequences} in many practical scenarios such as
Multiple Instance Learning (MIL). In the MIL scenario, a label is associated
with a set, instead of a single data instance. One of the application of MIL is
classification of high resolution histopathology images, called whole slide
images (WSIs). Each WSI is a gigapixel image with size $\approx$ 50,000 $\times$
50,000 pixels. The labels are generally associated with the entire WSI instead
of patch, region, or pixel level. MIL algorithms can be used to learn
representations of these WSIs by disassembling them into multiple representative
patches~\cite{KalraYottixelImageSearch2020,AdnanRepresentationLearningHistopathology,KalraPancancerdiagnosticconsensus2020}.

In this paper, we propose a novel architecture for exchangeable sequences
incorporating attention over the instances to learn inter-dependencies. We use
the results from Deep Sets~\cite{zaheer2017deep} to construct a permutation
invariant model for learning set representations. Our main contribution is a
sequence-to-sequence permutation invariant layer called \textbf{Memory
Block}. The proposed model uses a series of connected memory block layers, to
model complex dependencies within an input set using a self attention mechanism.
We validate our model using a toy datasets and two real-world applications. The
real world applications include, i) point cloud classification, and ii)
classification of WSI into two sub-type of lung cancers---Lung Adenocarcinoma
(LUAD)/ Lung Squamous Cell Carcinoma (LUSC) (see~\Autoref{fig:intro-luadlusc}).

The paper is structured as follows: \Autoref{sec:lit} discusses related and
recent works. We cover the mathematical concepts for exchangeable models in
\Autoref{sec:background}. We explain our approach and experimental results in
\Autoref{sec:method} and \Autoref{sec:results}.

\section{Related Work}
\label{sec:lit} In statistics, exchangeability has been long studied. de Finetti
studied exchangeable random variables and showed that sequence of infinite
exchangeable random variables can be factorised to independent and identically
distributed mixtures conditioned on some parameter $\theta$. Bayesian
sets~\cite{ghahramani2006bayesian} introduced a method to model exchangeable
sequences of binary random variables by analytically computing the integrals in
de Finetti's theorem. Orbanz et. al.~\cite{orbanz2014bayesian} used de Finetti's
theorem for Bayesian modelling of graphs, matrices, and other data that can be
modeled by random structures. Considerable work has also been done on partially
exchangeable random variables~\cite{aldous1981representations}.

Symmetry in neural networks was first proposed by Shawe et
al.~\cite{shawe1993symmetries} under the name \textit{Symmetry Network}. They
proposed that invariance can be achieved by weight-preserving automorphisms of a
neural network. Ravanbaksh et al. proposed a similar method for equivariance
network through parameter sharing~\cite{ravanbakhsh2017equivariance}. Bloem
Reddy et al.~\cite{bloem2019probabilistic} studied the concept of symmetry and
exchangeability for neural networks in detail and established similarity between
functional and probabilistic symmetry, and obtained generative functional
representations of joint and conditional probability distributions that are
invariant or equivariant under the action of a compact group. Zhou et.
al.~\cite{zhou2009multi} proposed treating instances in a set as non identical
and independent samples for multi instance problem.

Most of the work published in recent years have focused on ordered sets. Vinyals
et. al. introduced Order Matter: Sequence to Sequence for Sets in 2016 to learn
a sequence to sequence mapping. Many related models and key contributions have
been proposed that uses the idea of external memories like
RNNSearch~\cite{bahdanau2014neural}, Memory
Networks~\cite{weston2014memory,vaswani2017attention} and Neural Turing
Machines~\cite{graves2014neural}. Recent interest in exchangeable models was
developed due to their application in MIL. Deep Symmetry
Networks~\cite{gens2014deep} used kernel-based interpolation to tractably tie
parameters and pool over symmetry spaces of any dimension. Deep
Sets~\cite{zaheer2017deep} by Zaheer et al. proposed a permutation invariant
model. They proved that any pooling operation (mean, sum, max or similar) on
individual features is a universal approximator for any set function. They also
showed that any permutation invariant model follows de Finetti's theorem.Work
has also been done on learning point cloud classification which is an example of
MIL problem. Deep Learning with Sets and Point Cloud~\cite{ravanbakhsh2016deep}
used parameter sharing to get a equivariant layer. Another important paper on
exchangeable model is Set Transformer. Set
Transformer~\cite{DBLP:journals/corr/abs-1810-00825} by Lee et al. used results
from Zaheer et al.~\cite{zaheer2017deep} and proposed a
Transformer~\cite{vaswani2017attention} inspired permutation invariant neural
network. The Set Transformer uses attention mechanisms to attend to inputs in
order to invoke activation. Instead of using averaging over instances like in
Deep Sets, the Set Transformer uses a parametric aggregating function pool which
can adapt to the problem at hand. Another way to handle exchangeable data is to
modify RNNs to operate on exchangeable data. BRUNO \cite{korshunova2018bruno} is
a model for exchangeable data and makes use of deep features learned from
observations so as to model complex data types such as images. To achieve this,
they constructed a bijective mapping between random variables $x_i$ $\in X$ in
the observation space and features $z_i$ $\in Z$, and explicitly define an
exchangeable model for the sequences $z_1, z_2, z_3, \dots,z_n$. Deep Amortized
Clustering~\cite{lee2019deep} proposed using Set Transformers to cluster sets of
points with only few forward passes. Deep Set Prediction
Networks~\cite{zhang2019deep} introduced an interesting approach to predict sets
from a feature vector which is in contrast to predicting an output using sets.\\

\noindent $\textbf{MIL for Histopathology Image Analysis.}$
Exchangeable models are useful for histopathological images analysis as
ground-truth labeling is expensive and labels are available at WSI instead of at
the pixel level. A small pathology lab may process $\approx$10,000 WSIs/year,
producing a vast amount of data, presenting a unique opportunity for MIL
methods. Dismantling a WSI into smaller patches is a common practice; these
patches can be used for MIL. The authors in~\cite{IlseChapter22Deep2020} used
attention-based pooling to infer important patches for cancer classification. A
large amount of partially labeled data in histopathology can be used to discover
hidden patterns of clinical importance~\cite{KomuraMachineLearningMethods2018}.
Authors in~\cite{SudharshanMultipleinstancelearning2019} used MIL for breast
cancer classification. A permutation invariant operator introduced
by~\cite{TomczakHistopathologicalclassificationprecursor2018,TomczakDeepLearningPermutationinvariant2017}
was applied to pathology images. Recently, graph CNNs have been successfully used for
representation learning of
WSIs~\cite{AdnanRepresentationLearningHistopathology}. These compact and robust
representations of WSIs can be further used for various clinical applications
such as image-based search to make well-informed diagnostic
decisions~\cite{KalraPancancerdiagnosticconsensus2020,KalraYottixelImageSearch2020}.

\section{Background}
\label{sec:background}

This section explains the general concepts of exchangeability, its relation to
de~Finetti's theorem, and briefly discusses Memory Networks.\\

\noindent \textbf{Exchangeable Sequence. } A sequence of random variables
$x_1,\dots,x_n$ is exchangeable if the joint probability distribution
does not change on permutation of the elements in a set. Mathematically, if
$P(x_1,\dots,x_n) = P(x_{\pi(1)}, \dots , x_{\pi(n)})$ for a permutation
function $\pi$, then the sequence $x_1,\dots,x_n$ is exchangeable.\\

\noindent \textbf{Exchangeable Models.   } A model is said to be exchangeable if
the output of the model is invariant to the permutation of its inputs.
Exchangeability implies that the information provided by each instance $x_i$ is
independent of the order in which they are presented. Exchangeable models can
be of two types depending on the application: i) permutation invariant, and ii)
permutation equivariant.

A model represented by a function $f: X \to Y$ where $X$ is a set, is said to be
permutation equivariant if permutation of input instances permutes the output
labels with the same permutation $\pi$. Mathematically, a
permutation-equivariant model is represented as,
\begin{equation}
  \label{eq:perm-eq}
  f(x_{\pi(1)},x_{\pi(2)},\dots,x_{\pi(n)}) = [ y_{\pi(1)},y_{\pi(2)},\dots,y_{\pi(n)} ].
\end{equation}
Similarly, a function is permutation invariant if permutation of input instances
does not change the output of the model. Mathematically,
\begin{equation}
  \label{eq:perm-inv}
  f(x_1,x_2,\dots,x_n)= f(x_{\pi(1)},x_{\pi(2)},\dots,x_{\pi(n)}) .
\end{equation}

\noindent Deep Sets~\cite{zaheer2017deep} incorporate a permutation-invariant model to
learn arbitrary set functions by pooling in a latent space. The authors further
showed that any pooling operation such as averaging and max on individual
instances of a set can be used as a universal approximator for any arbitrary set
function. The authors proved the following two results about permutation invariant models.\\

\noindent \textbf{Theorem 1.  }A function $f(x)$ operating on a set
$X=\{x_1$,\dots,$x_n\}$ having elements from a countable universe, is a valid set
function, i.e., invariant to the permutation of instances in $X$, if it can be
decomposed to $\rho\left(\sum\phi(x)\right)$, for any function
$\phi$ and $\rho$.\\

\noindent \textbf{Theorem 2.  }Assume the elements are from a compact set in
$\mathbb{R}^d$, i.e., possibly uncountable, and the set size is fixed to $M$.
Then any continuous function operating on a set $X$, i.e., $f: \mathbb{R}^{d
\times M} \to \mathbb{R}$ which is permutation invariant to the elements in $X$
can be approximated arbitrarily close in the form of $\rho \sum(\phi(x))$.\\

The Theorem 1 is linked to de Finetti's theorem, which states that
a random infinitely exchangeable sequence can be factorised into
mixture densities conditioned on some parameter $\theta$ which captures the
underlying generative process i.e.
\begin{equation}
    P(x_1,\dots,x_n)=\int  p(\theta) \prod_{i=1}^{n} p(x_i|\theta)\ d(\theta).
\end{equation}

\noindent \textbf{Memory Networks. } The idea of using an external memory for
relational learning tasks was introduced by Weston et
al.~\cite{weston2014memory}. Later, an end-to-end trainable model was proposed
by Sukhbaatar et al.~\cite{sukhbaatar2015end}. Memory networks enable learning
of dependencies among instances of a set by providing an explicit memory
representation for each instance in the sequence. The idea of self attention is
popularized by~\cite{vaswani2017attention}, these models are known as
\emph{transformers}, widely used in NLP applications. The proposed MEM model
uses the self-attention (similar to transformers) within memory vectors,
aggregated using a pooling operation (weighted averaging) to form a
permutation-invariant representation (based on Theorems 1 and 2). The next
section explains it in details.

\section{Proposed Approach}
\label{sec:method}

\begin{figure}[t]
  \centering
    \begin{subfigure}[b]{0.55\textwidth}
    \centering
    \includegraphics[width=0.85\linewidth]{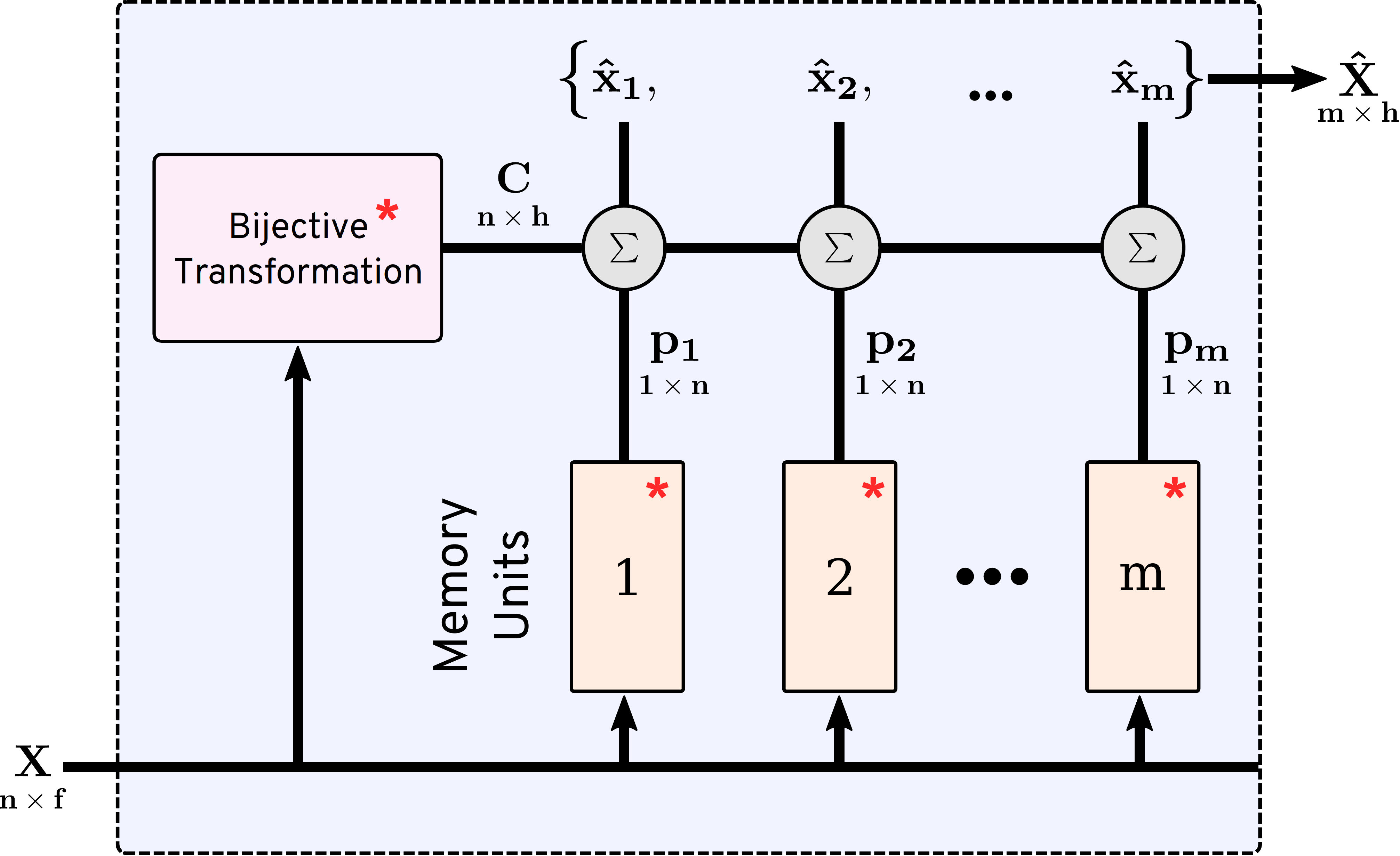}
    \caption{memory block}
    \label{fig:memblock}
  \end{subfigure}
  \begin{subfigure}[b]{0.43\textwidth}
    \centering
    \includegraphics[width=0.55\linewidth]{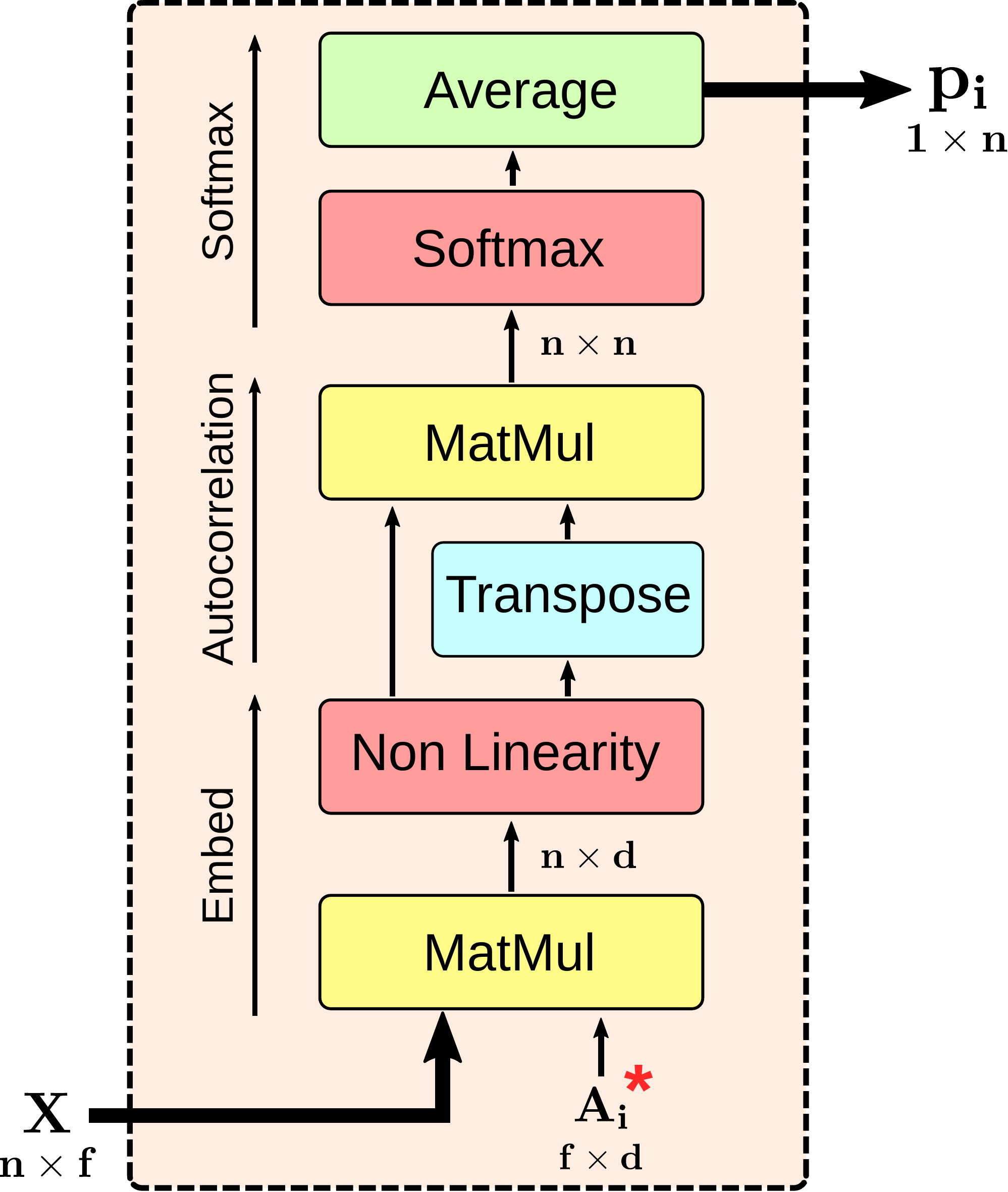}
    \caption{memory unit}
    \label{fig:memunit}
  \end{subfigure}
  \caption{ $X$ is an input sequence containing $n$ number of $f$-dimensional
    vectors. (a) The \textbf{memory block} is a sequence-to-sequence model that takes $X$
    and returns another sequence $\hat{X}$. The output
    $\hat{X}$ is a permutation-invariant representation of $X$. A bijective
    transformation model (an autoencoder) converts the input $X$ to a
    permutation-equivariant sequence $C$. The weighted sum of $C$ is computed over
    different probability distributions $p_i$ from memory units. The hyper-parameters of a memory block are
    i) dimensions of the bijective transformation $h$, and ii) number of memory units
    $m$. 
    (b) The \textbf{memory unit} has $A_i$, an embedding matrix (trainable parameters)
    that transforms elements of $X$ to a $d$-dimensional space (memories). The output $p_i$ is
    a probability distribution over the input $X$, also known as attention. The memory unit has a single 
    hyper-parameter $d$, i.e.~the dimension of the embedding
    space. ({\color{red}*} represents learnable parameters.) }
      \label{fig:comp}

\end{figure}

This section discusses the motivations, components, and offers an analysis of
the proposed \textbf{M}emory-based \textbf{E}xchangeable \textbf{M}odel
(MEM) capable of learning permutation invariant representation of sets and unordered sequences.

\subsection{Motivation}
In order to learn an efficient representation for a set of instances, it is
important to focus on instances which are ``important'' for a given task at
hand, i.e., we need to attend to specific instances more than other instances.
We therefore use the memory network to learn an attention mapping for each
instance. Memory networks are conventionally used for NLP for mapping questions
posted in natural language to an
answer~\cite{weston2014memory,sukhbaatar2015end}. We exploit the idea of having
\textit{memories} which can learn \textit{key} features shared by one or more
instances. Through these \textit{key} features, the model can learn
inter-dependencies using transformer style self-attention mechanism. As
inter-dependencies are learnt, a set can be condensed into a compact vector such
that a MLP can be used for a classification or regression learning.



\subsection{Model Components}
MEM is composed of four sequentially connected units: i) a feature extraction
model, ii) memory units, iii) memory blocks, and iv) fully connected layers to
predict the output.

A \textit{memory block} is the main component of MEM and learns a permutation
invariant representation of a given input sequence. Multiple memory blocks can
be stacked together for modeling complex relationships and dependencies in
exchangeable data. The memory block is made of memory units and a bijective
transformation unit shown in~\Autoref{fig:comp}\\

\noindent\textbf{Memory Unit.   }A memory unit transforms a given input sequence to an attention
vector. The higher attention value represents the higher ``importance'' of the
corresponding element of the input sequence. Essentially, it captures the
relationships among different elements of the input. Multiple memory
units enable the memory block to capture many complex dependencies and
relationships among the elements. Each memory unit consists of an embedding matrix $\mathbf{A_i}$ that transforms a $f$-dimensional input vector $x_j$ to a $d$-dimensional memory vector $u_{ij}$, as follows:
\[
  u_{ij} = \rho(x_j\mathbf{A_i}),
\]
where $\rho$ is some non-linearity. The memory vectors are stacked to form a
matrix $\mathbf{U_i} = [u_{i0},\dots,u_{in}]$ of the shape $(n \times d)$. The relative
degree of correlations among the memory vectors are computed using
cross-correlation followed by a column-wise softmax and then taking
a row-wise average, as follows:
\begin{equation}
  \label{eq:pi}
  \begin{split}
    S_{i} &= \text{column-wise-softmax}(\mathbf{U_i}\mathbf{U_i^T}), \\
    p_i &= \text{row-wise-average}(S_i),
  \end{split}
\end{equation}
The $p_i$ is the final output vector ($1 \times n$) from the $i^{th}$ memory unit $\mathbf{U_i}$, as shown in \Autoref{fig:comp}. The purpose of memory unit is to embed feature vectors into another space that
could correspond to a distinct ``attribute'' or ``characteristic'' of
instances. The cross correlation or the calculated attention vector represents
the instances which are highly suggestive of those ``attributes'' or
``characteristic''. We do not normalize memory vectors as magnitude of these
vectors may play an important role during the cross correlation.\\

\noindent\textbf{Memory Block. }A memory block is a sequence-to-sequence model,
i.e., it transforms a given input sequence $X = x_1,\dots,x_n$ to another
representative sequence $\hat{X} = \hat{x}_1,\dots,\hat{x}_m$. The output
sequence is invariant to the element-wise permutations of the input sequence. A
memory block contains $m$ number of memory units. In a memory block, each memory
unit takes a sequential data as an input and generates an attention vector.
These attention vectors are subsequently used to compute the final output
sequence. The schematic diagram of a memory block is shown
in~\autoref{fig:memblock}.\\

\noindent {The final output sequence $\hat{X}$} of a memory block is computed as
a weighted sum of $\mathbf{C}$ with the probability distributions
$p_1,\dots,p_m$ from all the $m$ memory units where $\mathbf{C}$ is a bijective
transformation of $X$ learned using an autoencoder. Each memory block has its
own autoencoder model to learn the bijective mapping. The $i^{th}$ element
$\hat{x}_i$ of the output sequence $\hat{X}$ is computed as matrix
multiplication of $p_i$ and $\mathbf{C}$, as follows:
\[
  \hat{x}_i = p_i \mathbf{C},
\]
where, $p_i$ is the output of $i^{th}$ memory unit given by~\eqref{eq:pi}.\\

\noindent {The bijective transformation} from $X \mapsto C$ enables
equivariant correspondence between the elements of the two sequences $X$ \&
$\hat{X}$, and maps two different elements in the input sequence to different
elements in the output sequence. It must be noted that bijective transformation
is permutation equivariant not invariant. The reconstruction maintains
one-to-one mapping between $X$ and $C$. The final output sequence from a memory
block is permutation invariant as it uses matrix multiplication between $p_i$
(attention) and $C$.


\begin{figure}[t]
  \centering 
  \includegraphics[width=0.99\linewidth]{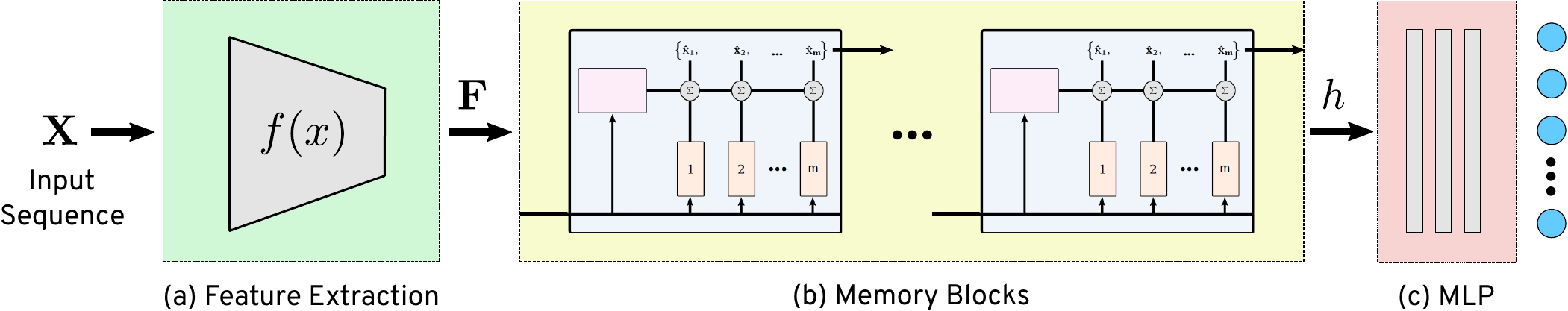}
  \caption{The overall architecture of the proposed Memory-based Exchangeable
    Model (MEM). The input to the model is a sequence, for e.g., a sequence of images or
    vectors. Each element of the input sequence $X$ is passed through \textbf{(a)}~feature
    extractor (CNN or MLP) to extract a sequence of feature vectors $F$, which is
    passed to \textbf{(c)}~sequentially connected memory blocks. A memory block outputs
    another sequence which is a permutation-invariant representation of the input
    sequence. The output from the last memory block is vectorized and given to
    \textbf{(c)}~MLP layers for classification/regression.}
  \label{fig:overallarch}
\end{figure}

\subsection{Model Architecture}
\begin{enumerate}
\item Each element of a given input sequence $X=x_1,\dots,x_n$ is passed through
  a feature extraction model to produce a sequence of feature vectors
  $F=f_1,\dots,f_n$. 
\item The feature sequence ${F}$ is then passed through a memory block to obtain
  another sequence $\hat{X}$ which is a permutation-invariant representation of the
  input sequence. The number of elements in the sequence $\hat{X}$ depends on the number of memory unit in the memory block layer.
\item Multiple memory blocks can be stacked in series. The output from the
  last memory block is either vectorized or pooled, which is subsequently passed
  to a MLP layer for classification or regression.
  
\end{enumerate}

\subsection{Analysis}

This section discusses the mathematical properties of our model. We use theorems from
Deep Sets~\cite{zaheer2017deep} to prove that our model is permutation invariant
and universal approximator for arbitrary set functions.\\

\noindent\textbf{Property 1.    } {Memory units are permutation equivariant.}

\noindent Consider an input sequence $X=x_1 \dots x_n$. Since, for each memory unit,
\[
\mathbf{U_i} = [\rho(x_o \mathbf{A_i}),\rho(x_1 \mathbf{A_i}),\dots,\rho(x_n \mathbf{A_i}) ] 
\]

\noindent By Equation~\eqref{eq:perm-eq}, $\mathbf{U_i}$ is permutation equivariant and thus $S_i$ in \eqref{eq:pi} is permutation equivariant. Finally, the attention vector $p_i$ is calculated by averaging all rows, therefore the final output of memory unit $p_i$ is permutation equivariant.\\


\noindent \textbf{Property 2.  } {Memory Blocks are permutation
invariant.} 

\noindent A memory block layer consisting of \textit{m} 
memory units generates a sequence $\hat{X} = \hat{x}_1,\dots,\hat{x}_m$ where $\hat{x_i}$ can be written as:
\[
\hat{x_i}= {p_i \mathbf{C}}
\]
\noindent Since both $\mathbf{C}$ and $p_i$ are permutation equivariant, 
therefore, $\hat{x_i}$, which is calculated by
matrix multiplication of $p_i$ and $\mathbf{C}$, is permutation invariant.\\



\section{Experiments}
\label{sec:results}

We performed two series of experiments comparing MEM against the simple pooling
operations proposed by Deep Sets ~\cite{zaheer2017deep}. In the first series of
experiments, we established the learning ability of the proposed model using toy
datasets. For the second series, we used two real-world dataset, i)
classification of subtypes of lung cancer against the largest public dataset of
histopathology whole slide images (WSIs)~\cite{weinstein2013cancer}, and ii) 3-D
object classification using Point Cloud Dataset~\cite{wu20153d}. \\

\noindent \textbf{Model Comparison. } We compared the performance of MEM against
Deep Sets~\cite{zaheer2017deep}. We use same the feature extractor for both Deep
Sets and MEM, and experimented with different choices of pooling
operations---max, mean, dot product, and sum. MEM also has a special pooling
``$mb1$'', which is a memory block with a single memory unit in the last hidden
layer. Therefore, we tested 9 different models for each experiment---five
configurations of our model, and four configurations of Deep Sets. We tried to
achieve the best performance by varying the hyper-parameters for each of the
configuration of both MEM and Deep Sets. We found that MEM had higher learning
capacity, therefore higher number of parameters resulted in better accuracy for
MEM but not necessarily for Deep Set. We denote the common feature extractor as
\textbf{FF} and Deep Sets as \textbf{DS} in the discussion below. The other
approaches that are compared have been appropriately cited.

\subsection{Toy Datasets}

\input{tab-toy-ds.tex}

To demonstrate the advantage of MEM over simple pooling operations, we consider
four toy problems, involving regression and classification over sets. We
constructed these toy datasets using the MNIST dataset.\\

\begin{figure}[t]
    \centering
    \includegraphics[width=0.55\linewidth]{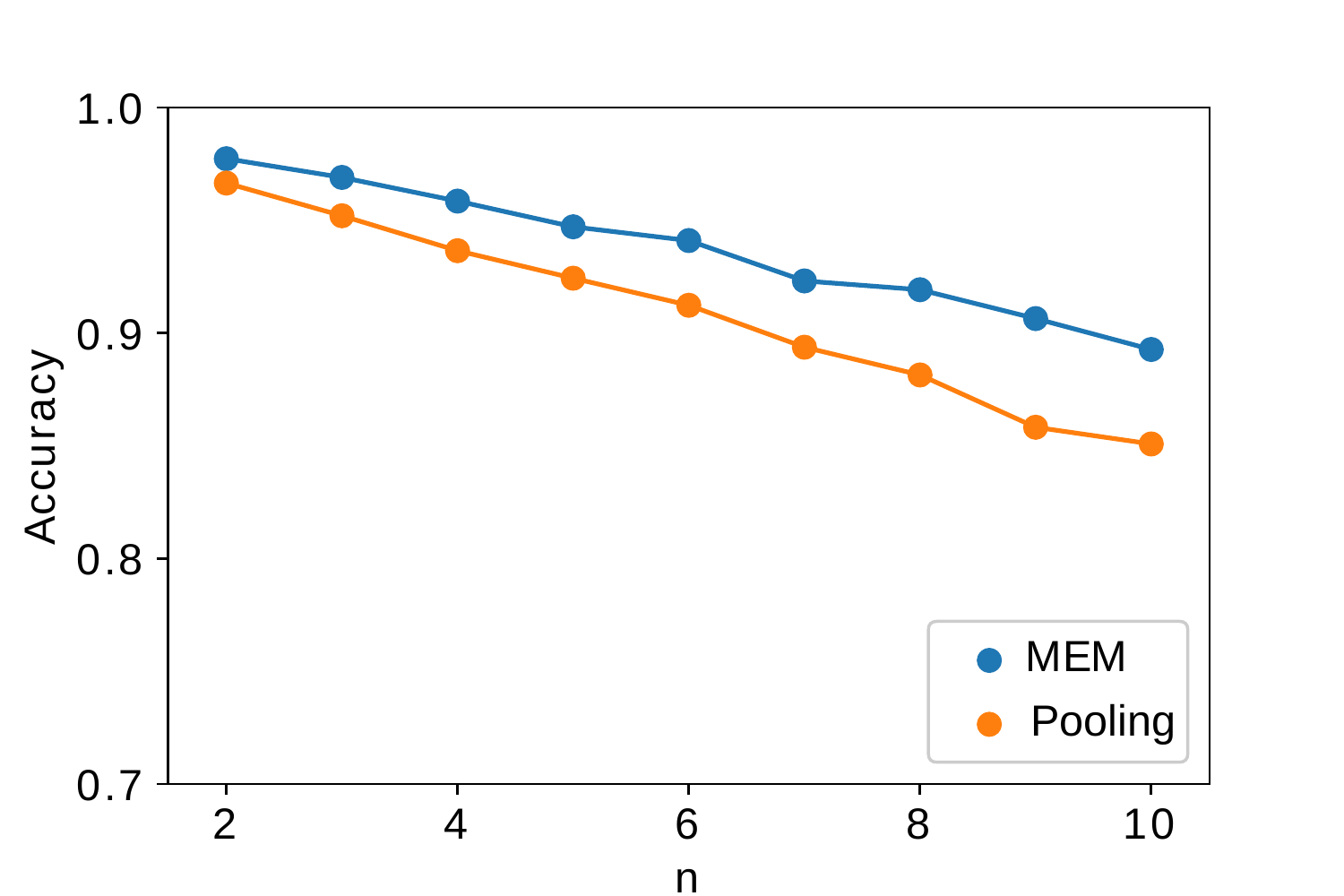}
    \caption{Comparison of MEM and feature pooling on a regression problem
      involving finding the sum of even digits within a set of MNIST images.
      Each point corresponds to the best configurations for the two models.}    
    \label{fig:sum-mem-ds}
\end{figure}

\noindent \textbf{Sum of Even Digits.} Sum of even digits is a regression problem over the set of
images containing handwritten digits from MNIST. For a given set of images $X =
\{x_1,\dots, x_n\}$, the goal is to find the sum of all even digits. We used the
Mean Absolute Error (MAE). We split the MNIST dataset into 70-30\%
training, and testing data-sets, respectively. We sampled 100,000 sets of 2 to
10 images from the training data. For testing, we sampled 10,000
sets of images containing $m$ number of images per set where $m \in [2, 10]$.
\autoref{fig:sum-mem-ds} shows the performance of MEM against simple pooling
operations with respect to the number of images in the set.\\

\noindent \textbf{Prime Sum.   } Prime Sum is a classification problem over a set of MNIST
images. A set is labeled positive if it contains any two digits such that their
sum is a prime number. We constructed the dataset by randomly sampling five
images from the MNIST dataset. We
constructed the training data with 20,000 sets randomly sampled from the
training data of MNIST. For testing, we randomly sampled 5,000 sets from the
testing data of MNIST. The
results are reported in the second column of~\autoref{tab:toy-ds-results} that
shows the robustness of memory block.\\

\noindent \textbf{Maximum of a Set.   }Maximum of a set is a regression problem to predict the
highest digit present in a set of images from MNIST. We constructed a set of
five images by randomly selecting samples from MNIST dataset. The label for each
set is the largest number present in the set. For example, images of $\{2, 5, 3,
3, 6\}$ is labeled as $6$. We constructed 20,000 training sets and for
testing we randomly sampled 5,000. The detailed comparison of accuracy and MAE between
different models is given in the second last column of
\autoref{tab:toy-ds-results}. We found that FF+Max learns the identity mapping
and thus results in a very high accuracy. In all the training sessions, we
consistently obtained the training accuracy of 100\% for the FF+Max configuration,
whereas MEM generalizes better than the Deep Sets.\\

\noindent \textbf{Counting Unique Images.   } Counting unique images is a regression problem over
a set. This task involves counting unique objects in a set of images from fashion
MNIST dataset~\cite{xiao2017fashion}. We constructed the training data by
selecting a set, as follows:
\begin{enumerate}
    \item Let n be the number of total images and u be the number of unique image in the set.
    \item Randomly select an integer $n$ between 2 and 10.
    \item Randomly select another integer $u$ between 1 and $n$.
    \item Select $u$ number of unique objects from fashion-MNIST training data.
    \item Then add $n$-$u$ number of randomly selected objects from the previous step. 
\end{enumerate}

\noindent The task is to count unique objects $u$ in a given set. The results are shown in the third column
of~\autoref{tab:toy-ds-results}.\\

\noindent \textbf{Amortized Gaussian Clustering.   } Amortized Gaussian clustering is a regression problem that
involves estimating the parameters of a population of Mixture of Gaussian (MoG).
Similar to Set Transformer~\cite{DBLP:journals/corr/abs-1810-00825}, we test our
model's ability to learn parameters of a Gaussian Mixture with $k$ components such that
the likelihood of the observed samples is maximum. This is in contrast to the EM
algorithm which updates parameters of the mixture recursively until the stopping
criterion is satisfied. Instead, we use MEM to directly predict parameters of
a MoG i.e. $f(x;\theta) = \{\pi(x), (\mu(x), \sigma(x))_{j=1}^{k}\}$. For simplicity we sample from MoG with only four components.
The Generative process for each training dataset is as follows

\begin{enumerate}
\item Mean of each Gaussian is selected from a uniform distribution i.e.
  $\mu_{j=1}^{k} \sim \text{Unif}(0,8)$.
\item Select a cluster for each instance in the set, i.e.,
  \[\pi \sim \text{Dir}([1,1]^{T}); z_{i} \sim \text{Categorical}(\pi)\]
\item Generate data from an univariate Gaussian $\sim
  \mathcal{N}(\mu_{z_{i}}, 0.3)$.
\end{enumerate}

\noindent We created a dataset of 20,000 sets each consisting of 500 points sampled from
different MoGs. Results in \autoref{tab:toy-ds-results} show that MEM is significantly better than Deep Sets.

\subsection{Real World Datasets}
To show the robustness and scalability of the model for the real-world problems,
we have validated MEM on two larger datasets. Firstly, we tested our model on a
point cloud dataset for predicting the object type from the set of 3D
coordinates. Secondly, we used the largest public repository of histopathology
images (TCGA)~\cite{weinstein2013cancer} to differentiate between two main
sub-types of lung cancer. Without any significant effort in extracting
histologically relevant features and fine-tuning, we achieved a remarkable
accuracy of 84.84\% on 5-fold validation.\\

\noindent \textbf{Point Cloud Classification.  } 
We evaluated MEM on a more complex classification task using
ModelNet40~\cite{wu20153d} point cloud dataset. The dataset consists of 40
different objects or classes embedded in a three dimensional space as points. We
produce point-clouds with 100 points (x, y, z-coordinates) each from the mesh
representation of objects using the point-cloud library’s sampling
routine~\cite{5980567}\footnote{We obtained the training and test datasets from
Zaheer et al.~\cite{zaheer2017deep}}. We compare the performance against various
other models reported in~\Autoref{tab:point_cloud}. We experimented with
different configurations of our model and found that FF+MB1 works best for 100
points cloud classification. We achieves the classification accuracy of 85.21\%
using 100 points. Our model performs better than Deep Sets and Set Transformer
for the same number of instances,
showing the effectiveness of having attention from memories.

\begin{table}[h]
  \centering
  \begin{tabular}{lcc}
    \Xhline{1.5\arrayrulewidth}
    Configuration                                              & Instance Size                & Accuracy        \\ \cline{1-2} 
    \hline
    3DShapeNet~\cite{wu20153d}                                 & 30$^3$                       & 0.77            \\
    Deep set~\cite{zaheer2017deep}                             & 100                          & 0.8200          \\
    VoxNet~\cite{MaturanaVoxNet3DConvolutional2015}            & 32$^2$                       & 0.8310          \\
    3D GAN~\cite{WuLearningProbabilisticLatent2017}            & 64$^3$                       & 0.833           \\
    Set Transformer~\cite{DBLP:journals/corr/abs-1810-00825}   & 100                          & 0.8454          \\
    Set Transformer~\cite{DBLP:journals/corr/abs-1810-00825}   & 1000                         & 0.8915          \\
    Deep set~\cite{zaheer2017deep}                             & 5000                         & 0.9             \\
    MVCNN~\cite{su15mvcnn}                                     & 164 $\times$ 164 $\times$ 12 & 0.901           \\
    Set Transformer~\cite{DBLP:journals/corr/abs-1810-00825}   & 5000                         & 0.9040          \\
    VRN Ensemble~\cite{BrockGenerativeDiscriminativeVoxel2016} & 32$^3$                       & 0.9554          \\
    \hline
    \textbf{FF + MEM + MB1 (Ours)}                             & \textbf{100}                 & \textbf{0.8521} \\
    \Xhline{1.5\arrayrulewidth}
  \end{tabular}
  \caption{Test accuracy for the point cloud classification on different
    instance sizes using various methods. MEM with configuration FF + MEM + MB1
    achieves 85.21\% accuracy for the instance size of 100 which is best
    compared to others. }
  \label{tab:point_cloud}
\end{table}

\noindent\textbf{Lung Cancer Subtype Classification. } 
Lung Adenocarcinoma (LUAD) and Lung Squamous Cell Carcinoma (LUSC) are two main
types of non-small cell lung cancer (NSCLC) that account for 65-70\% of all lung
cancers~\cite{graham2018classification}. Classifying patients accurately is
important for prognosis and therapy decisions. Automated classification of these
two main subtypes of NSCLC is a crucial step to build computerized decision
support and triaging systems. We present a two-staged method to differentiate
LUAD and LUSC for whole slide images, short WSIs, that are very large images.
Firstly, we implement a method to systematically sample patches/tiles from WSIs.
Next, we extract image features from these patches using
Densenet~\cite{huang2017densely}. We then use MEM to learn the representation of
a set of patches for each WSI.

\begin{figure}[t]
  \centering
  \begin{subfigure}[b]{0.45\textwidth}    
    \includegraphics[width=0.32\linewidth]{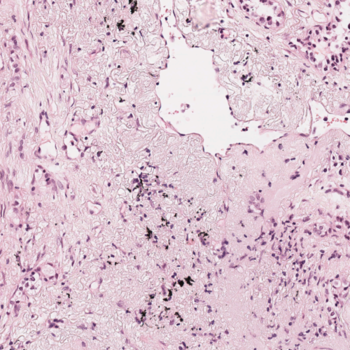}    
    \includegraphics[width=0.32\linewidth]{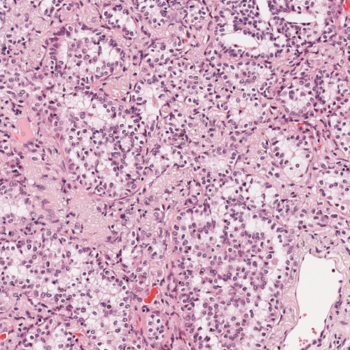}
    \includegraphics[width=0.32\linewidth]{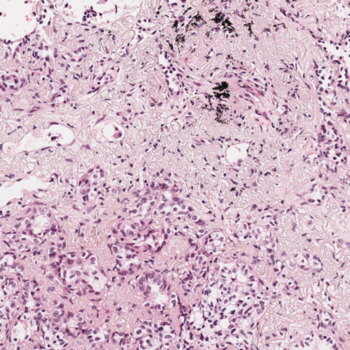}
    \includegraphics[width=0.32\linewidth]{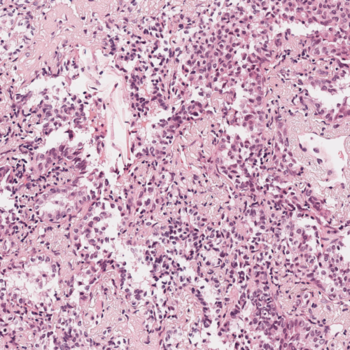}
    \includegraphics[width=0.32\linewidth]{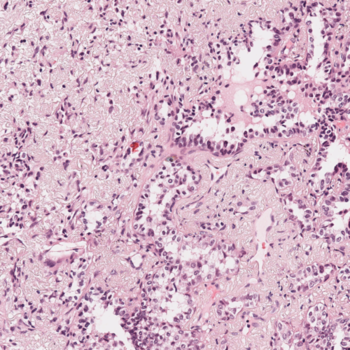}
    \includegraphics[width=0.32\linewidth]{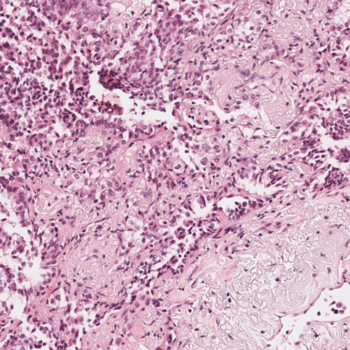}
    \caption{Lung Adenocarcinoma}
    \label{fig:1}
  \end{subfigure}
  \hspace*{.1cm}
  \begin{subfigure}[b]{0.45\textwidth}
    \includegraphics[width=0.32\linewidth]{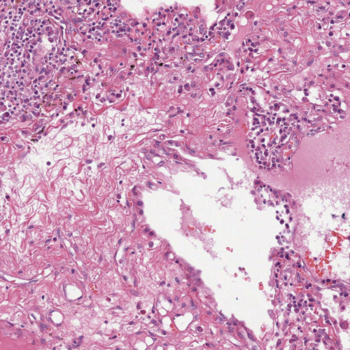}    
    \includegraphics[width=0.32\linewidth]{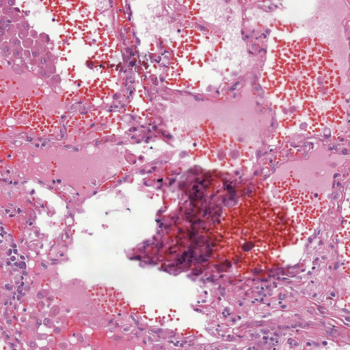}
    \includegraphics[width=0.32\linewidth]{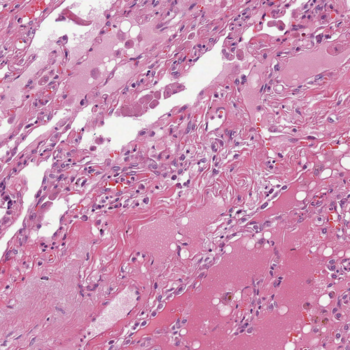}
    \includegraphics[width=0.32\linewidth]{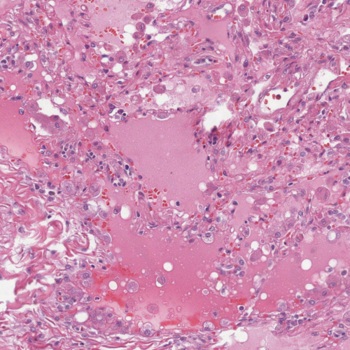}
    \includegraphics[width=0.32\linewidth]{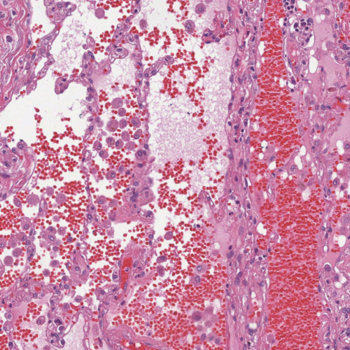}
    \includegraphics[width=0.32\linewidth]{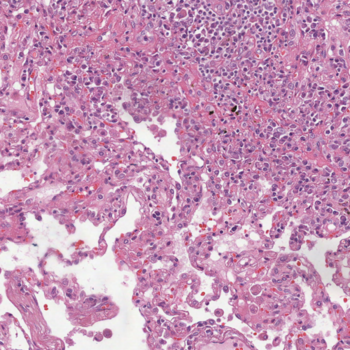}
    \caption{Lung Squamous Cell Carcinoma}
    \label{fig:2}
  \end{subfigure}
  \caption{The patches extracted from two WSIs of patients with (a) LUAD and (b)
  LUSC. Each slide roughly contains 500 patches.}
  \label{fig:luad-lusc-patches}
\end{figure}



To the best of our knowledge, this is the first ever study conducted on all the
lung cancer slides in TCGA dataset (comprising of 2 TB of data consisting of 2.5
million patches of size 1000$\times$1000 pixels). All research works in
literature use a subset of the WSIs with their own test-train split instead of
cross validation, making it difficult to compare against them. However, we have
achieved greater than or similar to all existing research works without
utilizing any expert's opinions (pathologists) or domain-specific techniques. We used 2,580 WSIs from TCGA public repository~\cite{weinstein2013cancer}
with 1,249, and 1,331 slides for LUAD and LUSC, respectively. We
process each WSI as follows.

\begin{enumerate}
\item \textbf{Tissue Extraction. } Every WSI contains a bright
  background that generally contains irrelevant (non-tissue) pixel information. We
  removed non-tissue regions using color thresholds.

\item \textbf{Selecting Representative Patches. } Segmented tissue is then
  divided into patches. All the patches are then grouped into a pre-set number of
  categories (classes) via a clustering method. A 10\% of all clustered patches
  are uniformly randomly selected distributed within each class to assemble
  \emph{representative patches}. Six of these representative patches for each
  class (LUAD and LUSC) is shown in~\autoref{fig:luad-lusc-patches}.
  
\item \textbf{Feature Set. } A set of features for each WSI is created by
  converting its representative patches into image features. We use
  DenseNet~\cite{huang2017densely} as the feature extraction model. There are a
  different number of feature vectors for each WSI.
\end{enumerate}

\begin{table}[ht]
  \centering
  \scalebox{0.9}{
  \begin{tabular}{lc}
    \Xhline{1.5\arrayrulewidth}
    Methods                  & Accuracy                     \\ \cline{1-2} 
    \hline
    \textbf{Coudray et al.~\cite{coudray2018classification}} & \textbf{0.85}          \\
    Jabber et al.~\cite{jaber2019automated}         & 0.8333 \\
    Khosravi et al.~\cite{khosravi2018deep}         & 0.83          \\
    Yu et al.~\cite{yu2016predicting}               & 0.75          \\
    \hline
    \textbf{FF + MEM + Sum (ours)} & \textbf{0.8484 $\pm$ 0.0210} \\
    FF + MEM + Mean (ours)         & 0.8465 $\pm$ 0.0225          \\
    FF + MEM + MB1 (ours)          & 0.8457 $\pm$ 0.0219          \\
    FF + MEM + Dotprod (ours)      & 0.6345 $\pm$ 0.0739          \\
    \hline
    FF + sum (DS)                     & 0.5159 $\pm$ 0.0120          \\
    FF + mean (DS)                      & 0.7777 $\pm$ 0.0273          \\
    FF + dotprod (DS)                   & 0.4112 $\pm$ 0.0121          \\
    \hline
    \Xhline{1.5\arrayrulewidth}
  \end{tabular}
  }
  \caption{Accuracy for LUAD vs LUSC classification for various methods. For our
  experiments, we conducted comprehensive 5-fold cross validation accuracy
  whereas other methods have used non-standardized test set. }
  \label{tab:lusc-luad}
\end{table}

The results are shown in~\autoref{tab:lusc-luad}. We achieved the maximum
accuracy of 84.84\% with FF + MEM + Sum configuration. It is difficult to
compare our approach against other approaches in literature due to
non-standardization of the dataset. Coudray et
al.~\cite{coudray2018classification} used the TCGA dataset with around 1,634
slides to classify LUAD and LUSC. They achieved AUC of 0.947 using patches at
20$\times$. We achieved a similar AUC of $0.94$ for one of the folds and average
AUC of $\mathbf{0.91}$. In fact, without any training they achieved the similar
accuracy as our model (around 85\%). It is important to note that we did not do any
fine-tuning or utilize any form of input from an expert/pathologist. Instead,
we extracted diverse patches and let the model learn to differentiate between
two sub-types by ``attending'' relevant ones. Another study by Jaber et
al.~\cite{jaber2019automated} uses cell density maps, achieving an accuracy of 83.33\% and AUC of 0.9068. However, they used much smaller portion of the
TCGA, i.e., 338 TCGA diagnostic WSIs (164 LUAD and 174 LUSC) were used to train,
and 150 (71 LUAD and 79 LUSC).

\section{Conclusions}
In this paper, we introduced a Memory-based Exchangeable Model (MEM) for learning
permutation invariant representations. The proposed method uses attention
mechanisms over ``memories'' (higher order features) for modelling complicated
interactions among elements of a set. Typically for MIL, instances are treated
as independently and identically distributed. However, instances are rarely
independent in real tasks, and we overcome this limitation using  an ``attention''
mechanism in memory units, that exploits relations among instances. We also
prove that the MEM is permutation invariant. We achieved good
performance on all problems that require exploiting instance relationships. Our
model scales well on real world problems as well, achieving an accuracy score of
84.84\% on classifying lung cancer sub-types on the largest public repository of
histopathology images.


\bibliographystyle{splncs04}
\bibliography{egbib}

\end{document}

%% file: tab-toy-ds.tex
\begin{table*}[]
  \centering
  \scalebox{0.45}{
  \begin{tabular}{l@{\hskip 0.05in}cc@{\hskip 0.15in}c@{\hskip 0.15in}cc@{\hskip 0.15in}cc@{\hskip 0.15in}c}
    \Xhline{1.5\arrayrulewidth}
                              & \multicolumn{2}{c@{\hskip 0.3in}}{Sum of Even} & Prime                        & \multicolumn{2}{c@{\hskip 0.3in}}{Counting Unique} & \multicolumn{2}{c@{\hskip 0.3in}}{Maximum of} & Gaussian                                                                                                                         \\
    Methods                   & \multicolumn{2}{c@{\hskip 0.3in}}{Digits}      & Sum                          & \multicolumn{2}{c@{\hskip 0.3in}}{Images}          & \multicolumn{2}{c@{\hskip 0.3in}}{Set}        & Clustering                                                                                                                       \\
                              & Accuracy                                       & MAE                          & Accuracy                                           & Accuracy                                      & MAE                           & Accuracy                              & MAE                                   & NLL \\ \hline
    FF + MEM + MB1 (ours)     & 0.9367 $\pm$ 0.0016                            & 0.2516 $\pm$ 0.0105          & 0.9438 $\pm$ 0.0043                                & 0.7108 $\pm$ 0.0084                           & 0.3931  $\pm$ 0.0080          & 0.9326 $\pm$ 0.0036                   & 0.1449 $\pm$ 0.0068                   & \textbf{1.348}    \\
    FF + MEM + Mean (ours)    & 0.9355 $\pm$ 0.0015                            & 0.2437 $\pm$ 0.0087          & 0.7208 $\pm$ 0.0217                                & 0.4264 $\pm$ 0.0062                           & 0.9525  $\pm$ 0.0109          & 0.9445 $\pm$ 0.0035                   & 0.1073 $\pm$ 0.0067
                              & 1.523                                                                                                                                                                                                                                                                                                      \\
    FF + MEM + Max (ours)     & \textbf{0.9431 $\pm$ 0.0020}                   & \textbf{0.2295 $\pm$ 0.0098} & 0.9361 $\pm$ 0.0060                                & 0.6888 $\pm$ 0.0066                           & 0.4140  $\pm$ 0.0079          & 0.9498 $\pm$ 0.0022                   & 0.1086 $\pm$ 0.0060                   & 1.388 \\
    FF + MEM + Dotprod (ours) & 0.8411 $\pm$ 0.0045                            & 0.3932 $\pm$ 0.0065          & \textbf{0.9450} $\pm$ \textbf{0.0086}              & \textbf{0.7284 $\pm$ 0.0055}                  & \textbf{0.3664  $\pm$ 0.0037} & \textbf{0.9517} $\pm$ \textbf{0.0041} & \textbf{0.0999} $\pm$ \textbf{0.0097} & 1.363 \\
    FF + MEM + Sum (ours)     & 0.9353 $\pm$ 0.0022                            & 0.2739 $\pm$ 0.0081          & 0.6652 $\pm$ 0.0389                                & 0.3138 $\pm$ 0.0094                           & 1.3696  $\pm$ 0.0151          & 0.9430 $\pm$ 0.0031                   & 0.1318 $\pm$ 0.0058                   & 1.611 \\ \hline
    FF + Mean (DS)            & 0.9159 $\pm$ 0.0019                            & 0.2958 $\pm$ 0.0049          & 0.5280  $\pm$ 0.0078                               & 0.3140 $\pm$ 0.0071                           & 1.2169  $\pm$ 0.0136          & 0.3223 $\pm$ 0.0075                   & 1.0029 $\pm$ 0.0155                   & 2.182 \\ 
    FF + Max (DS)             & 0.6291 $\pm$ 0.0047                            & 1.3292 $\pm$ 0.0211          & 0.9257  $\pm$ 0.0033                               & 0.7088 $\pm$ 0.0060                           & 0.3933  $\pm$ 0.0059          & \textbf{0.9585} $\pm$ \textbf{0.0012} & \textbf{0.0742} $\pm$ \textbf{0.0032} & 1.608 \\
    FF + Dotprod (DS)         & 0.1503 $\pm$ 0.0015                            & 1.8015 $\pm$ 0.0016          & 0.9224 $\pm$ 0.0028                                & \textbf{0.7254 $\pm$ 0.0063}                  & \textbf{0.3726  $\pm$ 0.0054} & \textbf{0.9548} $\pm$ \textbf{0.0017} & 0.1355 $\pm$ 0.0027                   & 8.538 \\
    FF + Sum (DS)             & 0.6333 $\pm$ 0.0043                            & 0.5763 $\pm$ 0.0069          & 0.5264   $\pm$  0.0050                             & 0.2982 $\pm$ 0.0042                           & 1.3415 $\pm$ 0.0169           & 0.3344 $\pm$  0.0038                  & 0.9645 $\pm$ 0.0111                   & 12.05 \\ 
    \Xhline{1.5\arrayrulewidth}
  \end{tabular}
  }
  \caption{Results on the toy datasets for different configurations of MEM and
    feature pooling. It must be noted that for Maximum of Set,  the
    configuration FF + Max (DS) achieves the best accuracy but it may predict
    the output perfectly by learning the identity function therefore we
    highlighted second best configuration FF + Dotprod (DS) as well.}
  \label{tab:toy-ds-results}
\end{table*}
